\pdfoutput=1

\documentclass[11pt]{article}
\usepackage{acl2014}
\usepackage{times}
\usepackage{amsmath}

\makeatletter
\newcommand{\@BIBLABEL}{\@emptybiblabel}
\newcommand{\@emptybiblabel}[1]{}
\makeatother
\usepackage[hidelinks]{hyperref}

\title{\textbf{GLEU Without Tuning}}
\newcommand{\clsp}{\ensuremath{{}^\text{1}}}
\newcommand{\hltcoe}{\ensuremath{{}^\text{2}}}
\newcommand{\yahoo}{\ensuremath{{}^\text{3}}}

\author{Courtney Napoles\clsp{\normalfont ,}
	Keisuke Sakaguchi\clsp{\normalfont ,}
	 Matt Post\hltcoe{\normalfont ,}
	\and Joel Tetreault\yahoo \\
	\clsp Center for Language and Speech Processing, Johns Hopkins University \\
	\hltcoe Human Language Technology Center of Excellence, Johns Hopkins University \\
	\yahoo Yahoo\\
}

\begin{document}
\maketitle

\begin{abstract}
    The GLEU metric was proposed for evaluating grammatical error corrections using n-gram overlap with a set of reference sentences, as opposed to precision/recall of specific annotated errors \cite{napoles-EtAl:2015:ACL-IJCNLP}.
    This paper describes improvements made to the GLEU metric that address problems that arise when using an increasing number of reference sets.
    Unlike the originally presented metric, the modified metric does not require tuning.
    We recommend that this version be used instead of the original version.\footnote{Download available at \url{https://github.com/cnap/gec-ranking/}.} 
\end{abstract}

\section{Introduction}
GLEU (Generalized Language Understanding Evaluation)\footnote{Not to be confused with the method of the same name presented in \newcite{mutton-EtAl:2007:ACLMain}.} was designed and developed using two sets of annotations as references, with a tunable weight to penalize n-grams that should have been changed in the system output but were left unchanged \cite{napoles-EtAl:2015:ACL-IJCNLP}.
After publication, it was observed that the weight needed to be re-tuned as the number of references changed. 
With more references, more variations of the sentence are seen which results in a larger set of reference n-grams.
Larger sets of reference n-grams tend to have higher overlap with the source n-grams, which decreases the number of n-grams that were seen in the source but not the reference.
Because of this, the penalty term decreases and a large weight is needed for the penalty term to have the same magnitude as the penalty when there are fewer references.

As re-tuning the weight for different sized reference sets is undesirable, we simplified GLEU so that there is no tuning needed and the metric is portable across comparisons against any number of references.

\section{Modifications to GLEU}
Our GLEU implementation differs from that of \newcite{napoles-EtAl:2015:ACL-IJCNLP}.
As originally presented, in computing n-gram precision, GLEU double-counts n-grams in the reference that do not appear in the source, and it subtracts a weighted count of n-grams that appear in the source ($S$) and not the reference ($R$).
We use a modified version, GLEU$^+$, that simplifies this. 
Precision is simply the number of reference n-gram matches, minus the counts of n-grams found more often in the source than the reference (Equation 1).
GLEU$^+$ follows the same intuition as the original GLEU: overlap between $S$ and $R$ should be rewarded and n-grams that should have been changed in $S$ but were not should be penalized.

\begin{figure*}[h]
	\centering
	\begin{align*}
	p^*_{n} =  \frac{
		\left( \sum\limits_{\textit{ngram}\in{\{C\cap R\}}} \textit{count}_{C,R}(\textit{ngram})
		- \sum\limits_{\textit{ngram}\in{\{C\cap S\}}}  \max\left[0,\textit{count}_{C,S}(\textit{ngram}) - \textit{count}_{C,R}(\textit{ngram})\right]
		\right)
	}
	{\sum\limits_{\textit{ngram}\in{\{C\}}}\textit{count}(\textit{ngram})}\\
	\textit{count}_{A,B}(\textit{ngram}) = \min\left( \textrm{\# occurrences of } \textit{ngram} \textrm{ in A}, \textrm{\# occurrences of }\textit{ngram}\textrm{ in B} \right) 
	\end{align*}
	Equation 1: Modified precision calculation of GLEU$^+$.
\end{figure*}

The precision term in Equation 1 is then used in the standard BLEU equation \cite{papineni-EtAl:2002:ACL} to get the GLEU$^+$ score.
Because the number of possible reference n-grams increases as more reference sets are used, we calculate an intermediate GLEU$^+$ by randomly sample from one of the references for each sentence, and report the mean score over 500 iterations.
It takes less than 30 seconds to evaluate 1,000 sentences using 500 iterations.

\section{Results}

Using this revised version of GLEU, we calculated the scores for each system submitted to the CoNLL 2014--Shared Task on Grammatical Error Correction\footnote{\url{http://www.comp.nus.edu.sg/~nlp/conll14st.html}} to update the results reported in Tables 4 and 5 of \newcite{napoles-EtAl:2015:ACL-IJCNLP}.
The system ranking by GLEU$^+$ is compared to the originally reported GLEU (GLEU$_0$), M$^2$, and the human ranking (Table \ref{tab-rank}).

\begin{table}
    \centering

    \begin{tabular}{|cccc|}
        \hline
        \rule{0pt}{2.5ex}    
    	\textbf{Human}& \textbf{M$^2$} & \textbf{GLEU$_0$} & \textbf{GLEU$^+$}\\
		\hline
        \rule{0pt}{2ex}    
		CAMB	& CUUI	  & CUUI      & CAMB\\
		AMU		& CAMB	  & AMU	      & CUUI\\
		RAC		& AMU	  & UFC	      & AMU\\
		CUUI	& POST	  & CAMB	  & UMC\\
		source	& UMC	  & source	  & PKU\\
		POST	& NTHU	  & IITB	  & POST\\
		UFC		& PKU     & SJTU	  & SJTU\\
		SJTU	& RAC     & PKU	      & NTHU\\
		IITB	& SJTU    & UMC	      & UFC\\
		PKU		& UFC     & NTHU	  & IITB\\
		UMC		& IPN	  & POST	  & source\\
		NTHU	& IITB    & RAC	      & RAC\\
		IPN		& source  & IPN	      & IPN\\
        \hline
    \end{tabular}
    \caption{\label{tab-rank} Ranking of CoNLL 2014 Shared Task system outputs, as judged by humans, M$^2$, and both versions of GLEU.}
\end{table}

On average, M$^2$ ranks systems within 3.4 places of the human ranking.
Both GLEU scores have closer rankings on average: GLEU$_0$ within 2.6 and GLEU$^+$ within 2.9 places of the human ranking.

The correlation between the system scores and the human ranking is shown in Table \ref{tab-corr}.
GLEU$^+$ has slightly stronger correlation with the human ranking than GLEU$_0$, which is significantly greater than the human correlation with M$^2$, however the rank correlation of GLEU$^+$ is weaker than GLEU$_0$ and M$^2$.

\section{Conclusion}
We recommend that the originally presented GLEU no longer be used due to the issues we identified in Section 1. 
The updated version of GLEU that does not require tuning (GLEU$^+$) should be used instead.
The code is available at\\ \url{https://github.com/cnap/gec-ranking}.

\newpage
\begin{table}[t]
    \centering
    \begin{tabular}{|c|cc|}
        \hline
        \rule{0pt}{2.5ex}
        \textbf{Metric} & $r$	 & $\rho$\\
        \hline
        \rule{0pt}{2.5ex}
        GLEU$^+$		& 0.549	 & 0.401\\
        GLEU$_0$		& 0.542	 & 0.555\\
        M$^2$			& 0.358	 & 0.429\\
        I-measure		& -0.051 & -0.005\\
        BLEU			& -0.125 & -0.225\\
        \hline
    \end{tabular}
    \caption{\label{tab-corr} Correlation between automatic metrics and the human ranking.}
\end{table}

\bibliographystyle{acl}
\bibliography{main}

\begin{thebibliography}{}

\bibitem[\protect\citename{Mutton \bgroup et al.\egroup
  }2007]{mutton-EtAl:2007:ACLMain}
Andrew Mutton, Mark Dras, Stephen Wan, and Robert Dale.
\newblock 2007.
\newblock Gleu: Automatic evaluation of sentence-level fluency.
\newblock In {\em Proceedings of the 45th Annual Meeting of the Association of
  Computational Linguistics}, pages 344--351, Prague, Czech Republic, June.
  Association for Computational Linguistics.

\bibitem[\protect\citename{Napoles \bgroup et al.\egroup
  }2015]{napoles-EtAl:2015:ACL-IJCNLP}
Courtney Napoles, Keisuke Sakaguchi, Matt Post, and Joel Tetreault.
\newblock 2015.
\newblock Ground truth for grammatical error correction metrics.
\newblock In {\em Proceedings of the 53rd Annual Meeting of the Association for
  Computational Linguistics and the 7th International Joint Conference on
  Natural Language Processing (Volume 2: Short Papers)}, pages 588--593,
  Beijing, China, July. Association for Computational Linguistics.

\bibitem[\protect\citename{Papineni \bgroup et al.\egroup
  }2002]{papineni-EtAl:2002:ACL}
Kishore Papineni, Salim Roukos, Todd Ward, and Wei-Jing Zhu.
\newblock 2002.
\newblock {BLEU}: A method for automatic evaluation of machine translation.
\newblock In {\em Proceedings of 40th Annual Meeting of the Association for
  Computational Linguistics}, pages 311--318, Philadelphia, Pennsylvania, USA,
  July. Association for Computational Linguistics.

\end{thebibliography}

\end{document}